\documentclass[10pt,twocolumn,letterpaper]{article}

\usepackage[pagenumbers]{cvpr} 

\usepackage[dvipsnames]{xcolor}

\definecolor{cvprblue}{rgb}{0.21,0.49,0.74}
\usepackage[pagebackref,breaklinks,colorlinks,citecolor=cvprblue]{hyperref}
\usepackage[accsupp]{axessibility}

\usepackage{graphicx}
\usepackage{amsmath}
\usepackage{amssymb}
\usepackage{booktabs}
\usepackage{times}
\usepackage{epsfig}
\usepackage{graphicx}
\usepackage{amsmath}
\usepackage{amssymb}
\usepackage{multirow}
\usepackage{pifont}
\usepackage{subcaption}
\usepackage[symbol]{footmisc}

\newcommand{\cmark}{\ding{51}}

\renewcommand{\arraystretch}{1.25}

\def\paperID{6} 
\def\confName{CVPR}
\def\confYear{2024}

\title{The devil is in discretization discrepancy. Robustifying Differentiable NAS with Single-Stage Searching Protocol}

\newcommand*\samethanks[1][\value{footnote}]{\footnotemark[#1]}
\author{Konstanty Subbotko\thanks{Work done while being at University of Warsaw}\qquad Wojciech Jablonski\samethanks \qquad Piotr Bilinski\\University of Warsaw}

\begin{document}
\maketitle
\begin{abstract}
Neural Architecture Search (NAS) has been widely adopted to design neural networks for various computer vision tasks. One of its most promising subdomains is differentiable NAS (DNAS), where the optimal architecture is found in a differentiable manner. However, gradient-based methods suffer from the discretization error, which can severely damage the process of obtaining the final architecture. In our work, we first study the risk of discretization error and show how it affects an unregularized supernet. Then, we present that penalizing high entropy, a common technique of architecture regularization, can hinder the supernet's performance. Therefore, to robustify the DNAS framework, we introduce a novel single-stage searching protocol, which is not reliant on decoding a continuous architecture. Our results demonstrate that this approach outperforms other DNAS methods by achieving $75.3\%$ in the searching stage on the Cityscapes validation dataset and attains performance $1.1\%$ higher than the optimal network of DCNAS on the non-dense search space comprising short connections. The entire training process takes only 5.5 GPU days due to the weight reuse, and yields a computationally efficient architecture. Additionally, we propose a new dataset split procedure, which substantially improves results and prevents architecture degeneration in DARTS.
\end{abstract}    
\section{Introduction}
\label{section:introduction}

Neural architecture search (NAS) is a field that automates the designing of neural networks. Differentiable neural architecture search (DNAS) denotes the set of gradient-based NAS techniques. In these methods, we relax the discrete architecture space into the space of continuous architectures~\cite{darts} and optimize it using stochastic gradient descent.

DNAS framework can be decomposed into three stages:
\begin{enumerate}
    \item the \textit{searching} stage, where a ``supernet'' assembling all architecture candidates as subnetworks is trained,
    \item the \textit{decoding} stage, which retrieves a discrete architecture from a continuous search space, and
    \item the \textit{retraining} stage, where a retrieved architecture is trained for a longer time and with newly initialized weights.
\end{enumerate}

\begin{table}
\begin{center}
\begin{tabular}{|c|c||c|c|}
\hline
\multirow{2}{*}{\textbf{Method}} & \multirow{2}{*}{\textbf{Parameters}} & \multicolumn{2}{c|}{\textbf{Entropy}} \\\cline{3-4} &  & \textbf{Start} & \textbf{End}\\
\hline\hline
\multirow{2}{*}{Auto-DeepLab~\cite{autodeeplab}} & Edges & 0.259 & 0.256\\
 & Operations & 0.260 & 0.258\\
 \hline
\multirow{2}{*}{DARTS~\cite{darts}} & Topology & 0.126 & 0.126\\
& Operations & 0.260 & 0.256\\
\hline
\multirow{2}{*}{Ours} & Edges & 0.128 & 0.127\\
& Operations & 0.346 & 0.345\\
\hline
\end{tabular}
\end{center}
\caption{Lack of implicit entropy regularization in a supernet. The average entropy of architectural parameters across different methods and different tasks at the start and the end of the searching stage. For DARTS, we adopt decoupled topology search~\cite{dots}, which introduces a new set of parameters.}
\label{table:dnas_entropy}
\end{table}

\begin{figure*}
\begin{center}
\includegraphics[width=\linewidth,]{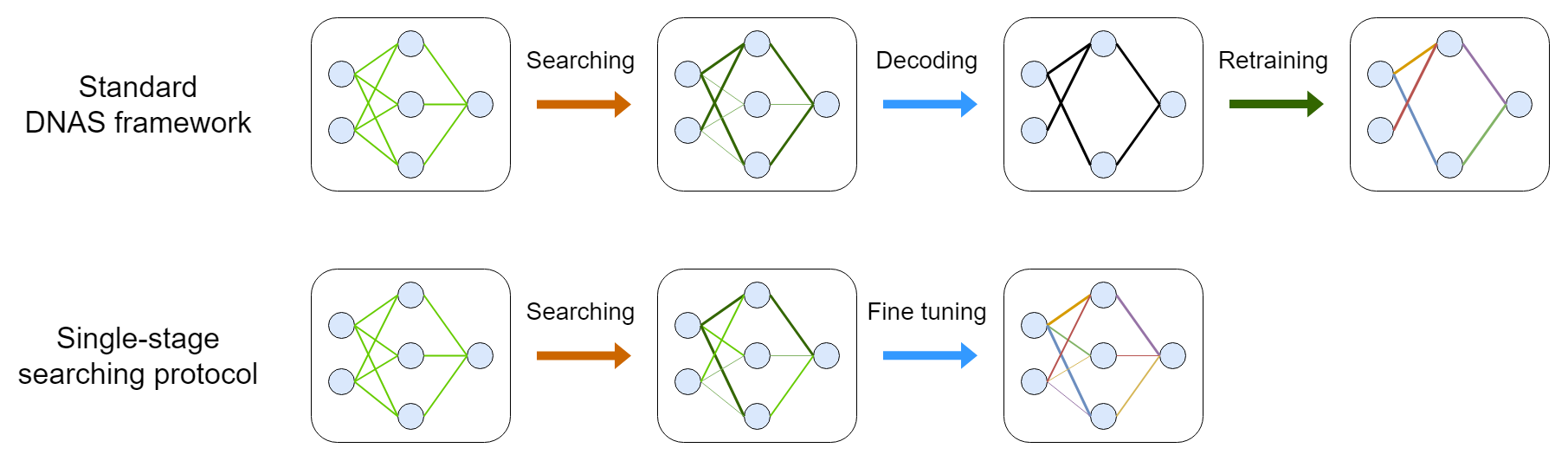}
\end{center}
\caption{Illustration of the single-stage searching protocol. We replace both the decoding and the retraining stages with a new fine-tuning phase, during which architecture is frozen. By reusing weights, we save a considerable amount of the retraining time. We keep the optimized architectural parameters in the final network, which means that edges in a supernet take on real values, unlike in the standard DNAS framework.}
\label{fig:comparison}
\end{figure*}

The usage of supernet greatly reduces computational costs by enabling weight-sharing across a vast number of different architectures~\cite{darts,enas}. However, despite its computational effectiveness and significant potential, practical applications of DNAS are hindered by the severe fragility and instability~\cite{robustdarts,darts+,fairdarts}. One of the major issues, which we refer to as the \textit{discretization error}, concerns the poor architecture optimization process and emerges at the decoding stage during the discretization procedure~\cite{fairdarts,da2s,goldnas,robustdarts}. Discarding operations or connections can yield substantially different architecture when a supernet is poorly discretized and has a high entropy at the end of the training. As a result, it can impact the searching-retraining correlation. Thus, even a network retrieved from a well-performing supernet might underperform after retraining.

In this work, we shed more light on the discretization error and propose a novel solution to address it. Contrary to the common approach, we perform experiments in the semantic segmentation task~\cite{cityscapes} for two reasons. First, our approach is better suited for tasks that can benefit from dense architectures and a certain design of the search space. Second, we find it more effective in highlighting some problems than the extensively studied task of image classification, where differences between approaches can even be statistically insignificant. Our experiments in \cref{table:dnas_entropy} reveal a lack of implicit entropy regularization in the vanilla DNAS framework. We can observe that the average entropy of architectural parameters remains constant throughout the training, indicating that a considerable number of operations in a supernet contributes to the prediction at the end of the searching. This might in turn cause the discretization error.

One of the common approaches to tackle the discretization error is to impose a one-hot distribution over architectural parameters by regularizing a supernet~\cite{fairdarts,dcnas,da2s}. In our work, we take a closer look at the method of penalizing high entropy and highlight its shortcomings. As we demonstrate in \cref{sec:experiments}, such a regularization induces a trade-off between discretization and obtained results. Namely, we show that the magnitude of the entropy loss negatively correlates with the discretization error, which indeed suggests a need for a strong regularization. At the same time, our experiments indicate that it can degrade supernet's performance in the searching stage. We also consider another variant of dynamic entropy loss regularization~\cite{da2s}, which alleviates the performance issue, but does not eliminate it entirely.

To this end, we propose to approach the problem from a different angle and to train the supernet in a fully proxyless manner by introducing a \textit{single-stage} searching protocol. The approach is illustrated in \cref{fig:comparison}. We simplify the searching process by replacing both the decoding and the retraining stages with a new fine-tuning phase on the top of the searching stage. We do not perform discretization and, thereby, we treat optimized supernet with all its trained parameters as the final model. By reusing weights from the searching stage, we considerably reduce the total training time. In this approach, it is crucial to design a search space that is both expressive and computationally efficient. For that reason, our method might not be yet appropriate for certain tasks. We apply our single-stage searching method to the non-dense DCNAS search space \cite{dcnas}, which includes transmissions spanning only between consecutive layers. As we show in \cref{sec:experiments}, our approach is on par with, or even surpasses, other state-of-the-art DNAS models in terms of computational requirements.

The single-stage searching protocol also addresses another deficiency of the DNAS framework, which is often overlooked - prohibitively high computational complexity. Because of the proxy searching procedure, each retrieved architecture is trained from scratch, which imposes extra costs. Furthermore, the DNAS method can suffer from a poor searching-retraining correlation~\cite{dcnas}. As a result, in addition to the costs of finding the optimal network, several candidate architectures must be evaluated to counteract low correlation. Instead, our single-stage searching algorithm takes only 5.5 GPU days to converge by using a single set of weights throughout the training.

We validate our improvements on the Cityscapes dataset~\cite{cityscapes}. 
Our single-stage method outperforms other DNAS methods in the searching stage by achieving $75.3\%$ on the validation set. Moreover, it attains performance $1.1\%$ higher than the final derived network of DCNAS on the non-dense search space, which demonstrates the viability of the single-stage approach. In our experiments, we also perform an ablation study on a dataset split procedure~\cite{darts,autodeeplab}, which shows that more optimal data usage yields up to $5.4\%$ boost. Additionally, we demonstrate that it can prevent architecture degeneration in DARTS.

We summarize our main contributions as follows:
\begin{itemize}
    \item We investigate the discretization error in a semantic segmentation task, empirically show its negative consequences, depending on the strength of the regularization.
    \item We study the entropy architecture regularization and demonstrate that it hinders the supernet's performance.
    \item We introduce a fully proxyless, single-stage searching protocol that eliminates the discretization error by thoroughly fine-tuning the supernet. We demonstrate that it yields a computationally efficient architecture, which outperforms DCNAS on a comparable search space. Also, we show its superiority in terms of the total training time.
    \item We conduct an ablation study on the training dataset split approach in DNAS methods, highlighting a much more optimal dataset usage, which considerably enhances performance and prevents architecture degeneration in DARTS.
\end{itemize}

\section{Related work}\label{section:related_work}

\textbf{Neural architecture search} is a collection of novel techniques aiming to automate the neural network design process. It can automatically discover the optimal neural network architecture
for a given task or dataset. NAS research concerns primarily evolutionary \cite{nas_evol_2,nas_evol_1,nas_evol_3}, reinforcement learning \cite{nas_rl_2,nas_rl_4,nas_rl_3,nas_rl_1,nas_rl_5} and DNAS \cite{darts,dnas_1,dnas_2,fbnet,fbnet_v2} methods.

In Differentiable NAS (DNAS), the optimal architecture can be found using stochastic gradient descent, thanks to the differentiable representation of the search space~\cite{darts}. Each architectural choice is assigned a continuous parameter. The search space is represented as a large, weight-sharing supernet, where different architecture candidates correspond to different subsets of this supernet. Searching over this search space essentially comes down to training the network. Afterward, the optimal discrete architecture is retrieved from a supernet and retrained from scratch for a longer time.

NAS algorithms search for a network within a predefined search space. The search space must be expressive enough to include a wide range of candidate architecture and also be efficiently optimizable. In NASNet \cite{nas_rl_5}, the network comprises a sequence of convolutional cells, all sharing the same architecture. The cell, composed of multiple blocks, forms the search space. The spatial dimension throughout the network is controlled manually. Several NAS applications to the image classification task follow the same design~\cite{pnasnet,darts,nas_rl_4,nas_rl_5}.

AutoDeepLab \cite{autodeeplab} builds upon DARTS \cite{darts} and adapts its approach to semantic segmentation by introducing a significantly larger network-level hierarchical architecture search space, while operating on the same cell-level search space as DARTS. DCNAS \cite{dcnas} extends the idea of hierarchical search space and introduces a densely connected search space, incorporating long-range connections reaching every cell and every spatial level. DPC~\cite{dpc} proposes a recursive search space formed by a dense prediction cell, which uses a segmentation-specific set of operations, such as atrous convolution or pooling. In the searching stage, DPC constructs a proxy task of finding the dense prediction cell on the top of the backbone pretrained on ImageNet~\cite{imagenet}.

\textbf{Discretization issue.} Several works derived from DARTS~\cite{darts} focus on the searching to retraining transition. Much attention was drawn to the collapsing phenomenon in DARTS, where a supernet assigns excessive weights to skip connections~\cite{robustdarts,fairdarts,darts+,smoothdarts,darts-,betadarts}. RobustDARTS~\cite{robustdarts} alleviates this issue through a hand-crafted early-stopping strategy. SmoothDARTS~\cite{smoothdarts} enhances architecture generalization by perturbing architecture parameters, thus smoothing the loss landscape. Some other works improve DARTS decoding efficacy by better estimating the importance of operations~\cite{dartspt,influencefun}.

The concept of discretization issue is established in the literature~\cite{fairdarts,goldnas,da2s,robustdarts}. Fair DARTS~\cite{fairdarts} observes the discretization discrepancy by visualizing softmax activations and introduces a zero-one loss, which imposes a one-hot distribution. GOLD-NAS~\cite{goldnas} addresses the issue by using hardware constraints to penalize significant architectural parameters and progressively prune weak operations. The most related to our work in terms of discretization study is DA$^2$S~\cite{da2s}. The authors show that vanilla DARTS suffers from a performance collapse by performing inference on a discretized supernet. To alleviate this collapse, they introduce dynamic entropy loss to impose one-hot distribution in the later stages of the training. DCNAS~\cite{dcnas} likewise regularizes architectural parameters to diminish insignificant transmissions.

\begin{figure*}
\centering
\begin{subfigure}{0.475\textwidth}
\centering\includegraphics[width=\linewidth,]{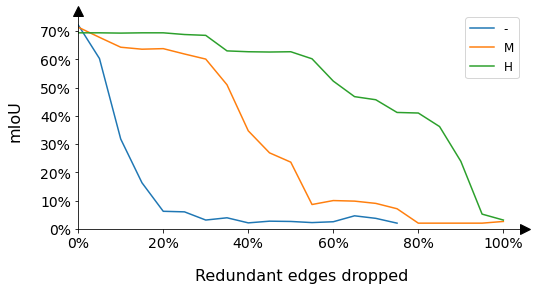}
\caption{Discretiation w/o fine-tuning}
\label{fig:inference}
\end{subfigure}
\hspace{\fill}
\begin{subfigure}{0.475\textwidth}
\centering
\includegraphics[width=\linewidth,]{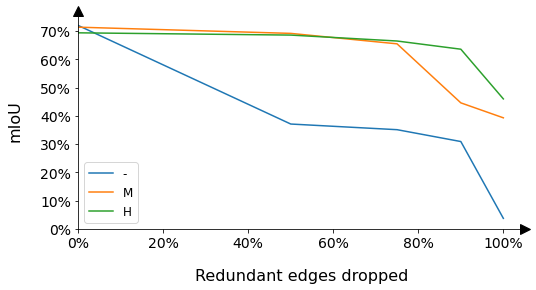}
\caption{Discretiation with fine-tuning}
\label{fig:inference_ft}
\end{subfigure}
\caption{Visualization of the discretization error across different entropy loss magnitudes. For more details, see \cref{subsection:discretization_error}.}
\end{figure*}

\textbf{Proxyless searching} involves sharing the training protocol between a supernet and a retrained network. In particular, this implies using the same hyperparameters, such as batch size or image crop. ProxylessNAS~\cite{proxylessnas} samples paths within the search space to facilitate proxyless searching. Similarly, DCNAS~\cite{dcnas} probes candidate architectures by sampling connections. This approach, along with masking a subset of channels in each cell~\cite{dcnas,pcdarts}, makes proxyless searching viable. Another study~\cite{progressivedarts} divides the searching stage into a few phases and performs a stepwise discretization. This can be considered a related work to proxyless searching, as the proxy gradually decreases. In our work, we extend the idea of proxyless searching and propose to optimize a supernet in an end-to-end manner using target hyparameters. However, unlike other approaches, we do not retrain it. Instead, we reuse the already trained weights, thus significantly reducing computational requirements.
\section{Methods}\label{sec:methods}

In this section, we first present cell-level and network-level architecture search space. Subsequently, we introduce the single-stage searching protocol, which can save computational costs and eliminate the discretization error, followed by a description of the entropy loss.

\subsection{Search space}

\textbf{Network-level architecture.} The supernet comprises three modules: the \textit{stem}, the \textit{backbone}, and the \textit{decoder}. Following DCNAS \cite{dcnas}, we utilize multi-scale feature representation in our network. We adopt the stem module used by Auto-DeepLab and adjust it to the multi-scale network structure by performing interpolations of an input image. For the decoder, we reuse the prediction head designed by DCNAS.

A dense rectangular-like grid of cells forms the backbone. Each cell corresponds to a particular layer and resolution. Resolutions reach up to a downsampling rate of 32. Network-level transmissions span between adjacent cells in consecutive layers. We assign an architectural parameter $\beta_{s' \rightarrow s}^l$ to a transition in layer $l$ between resolution $s'$ and $s$. We define an input to a cell as a weighted average over the outputs of its predecessors:

\begin{equation}\label{eq:1}
X_s^l = \sum_{t\hspace{0.5mm}\in\{s/2,\hspace{0.4mm}s,\hspace{0.4mm}2s\}} \mathcal{P}  \big( Y_{t}^{l-1} \big) \hspace{1mm} \hat\beta_{t \rightarrow s}^{l} \hspace{0.5mm}.
\end{equation}

\noindent
Here, $X_j^i$ and $Y_j^i$ denote the input and output of an $j$-th cell in a $i$-th layer, respectively. $\mathcal{P}$ is a shape-aligning preprocessing operation applied separately to each feature-maps.
$\hat\beta$ are normalized scalars, indicating edge relative importance within a cell:

\begin{equation}\label{eq:2}
\hat \beta_{s' \rightarrow s}^l = \frac{\exp(\beta_{s' \rightarrow s}^l)}{\sum_{t\hspace{0.5mm}\in\{s/2,\hspace{0.4mm}s,\hspace{0.4mm}2s\}}\exp(\beta_{t \rightarrow s}^l)}\hspace{0.5mm}.
\end{equation}

\textbf{Cell-level architecture.} We follow DCNAS and use an inverted bottleneck \cite{mobilenetv3} cell structure similar to \cite{proxylessnas}. Operator space consists of convolutions with different kernel sizes. We assign a parameter $\alpha_{s, l}^k$ to a block with convolution $o_k$ with a kernel of size $k$ in each cell. The output of a cell is defined as follows:

\begin{equation}\label{eq:1}
Y_s^l = \sum_k\hat\alpha_{s, l}^k \hspace{1mm} o_k \big(X_s^l\big) \hspace{0.5mm},
\end{equation}

\noindent
where $\hat\alpha_{s, l}^k$ are parameters normalized using softmax, analogously to network-level transmissions. In certain experiments, we adopt the channel sampling scheme \cite{pcdarts,dcnas}, which reduces the number of feature-maps filters before processing them in a cell.

\subsection{Single-stage searching protocol}

We propose to train the network in a fully proxyless way by introducing the single-stage searching protocol. We shrink the DNAS framework by dropping the decoding and retraining stages, retaining only the searching stage, which comprises three phases:

\begin{enumerate}
    \item \textit{warmup} phase, which precedes architecture optimization, and with gradient updates performed exclusively on weights,
    \item \textit{searching} phase, where architecture and weights are jointly optimized,
    \item \textit{fine-tuning}, in which architecture is fixed, and only the weights are optimized.
\end{enumerate}

The warmup phase has been introduced to prevent architecture from degeneration caused by using randomly initialized weights~\cite{autodeeplab}. In the searching phase, we alternately update architectural parameters and weights by adopting the following bilevel optimization scheme:

\vbox{
    \begin{itemize}
        \item Update network weights $w$ by $\nabla_w\mathcal{L}_A(w, \alpha, \beta)$,
        \item Update architecture $\alpha, \beta$ by $\nabla_{\alpha, \beta}\mathcal{L}_B(w, \alpha, \beta)$,
    \end{itemize}
}

\noindent
where $\mathcal{L}_A$ and $\mathcal{L}_B$ denote cross-entropy loss computed on two subsets of training data $A$ and $B$, respectively. We also include the entropy regularization term in $\mathcal{L}_B$.

The fine-tuning phase can be perceived as a replacement for the retraining stage. However, we managed to reduce the overall training time considerably. Unlike in the standard DNAS framework, we do not perform decoding, and thus, we can efficiently reuse weights that have already been trained in the preceding phases. In \cref{sec:experiments}, we show that this approach can achieve optimal results.

\subsection{Entropy loss}\label{subsec:entropy_loss}

\begin{figure}
\begin{center}
\includegraphics[width=\linewidth,]{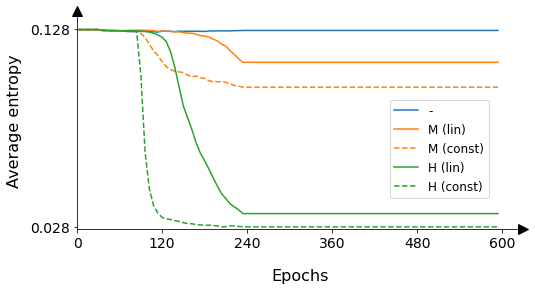}
\end{center}
\caption{The average entropy of architectural parameters throughout the training. Dashed and solid curves correspond to supernets trained with the constant and the linear entropy scaling function, as described in \cref{subsec:entropy_loss}. Curves denoted by -, M, and H refer to supernets trained without entropy loss, with medium entropy loss, and with high entropy loss, respectively.}
\label{fig:entropy}
\end{figure}

Similar to other works \cite{da2s,dcnas}, we use the entropy loss term in our experiments to regularize architectural parameters. The purpose is to penalize the excessive usage of insignificant transmissions and operations in a supernet. We formulate the term in the following way:

\vspace{1mm}
\begin{equation}\label{eq:5}
\mathcal{L}_\text{ent}\hspace{1mm}=\hspace{1mm}c_\beta\hspace{2mm}f(t)\hspace{1mm}\sum_{\hat\beta_i}\hspace{1mm}\hat\beta_i\hspace{1mm}\ln{\hat\beta_i}\hspace{0.5mm}.
\end{equation}
\vspace{1mm}

\noindent
Here, $f(t)$ is a scaling term dependent on the time \textit{t}, and $c_\beta$ denotes the overall entropy loss magnitude for the network-level architectural parameters. We apply the same regularization to the cell-level parameters $c_\alpha$.

In our experiments, we consider two variants of the scaling function. The \textit{Linear} function refers to a linear scaling term, which gradually increases the magnitude of the entropy loss from $0$ to $1$. The \textit{Constant} function sets $f(t) = 1$ and corresponds to the default approach of applying the entropy loss.

\setlength{\tabcolsep}{5pt}
\renewcommand{\arraystretch}{1.3}

\begin{table*}
\begin{center}
\begin{tabular}{|c||cccc|}
\hline
\multirow{2}{*}{\textbf{Function}} & \multicolumn{4}{c|}{\textbf{Entropy magnitude}} \\
  & - & L & M & H \\
\hline\hline
Constant & $\pmb{70.9} \pm 1.1\%$ & $70.2 \pm 1.7\%$ & $69.6 \pm 0.9 \%$ & $68.1 \pm 1.2\%$ \\
Linear & $\pmb{70.9} \pm 1.1\%$ & $\pmb{70.9} \pm 0.4\%$ & $70.8 \pm 0.7\%$ & $68.8 \pm 0.5\%$ \\
\hline
\end{tabular}
\end{center}
\caption{Results for different scaling functions and magnitudes averaged over three runs. \textit{Function} denotes the time-dependent scaling term introduced in \cref{subsec:entropy_loss}. \textit{Entropy magnitude} refers to the different values of $c_\alpha$ and $c_\beta$. \textit{L}, \textit{M}, \textit{H} denote low, medium and large magnitudes, respectively.}
\label{table:entropy}
\end{table*}
\section{Experiments}\label{sec:experiments}

We validate our methods on Cityscapes \cite{cityscapes}, a go-to semantic segmentation dataset for evaluating the searching efficacy of DNAS. It enables us to directly compare our approach with other methods in a challenging environment. The dataset consists of high-resolution images with fine and coarse annotations. The former are split into sets of 2975, 500, and 1525 images for training, validation, and testing, respectively. The latter provides labels for 20000 training samples, but details and object boundaries are coarsely annotated.

\subsection{Implementation details}\label{subsec:implementation}

The large model takes 1.4 days to train for 600 epochs on 4 Tesla V100 32GB GPUs. We use syncBN \cite{syncbn} to synchronize statistics in the batch normalization layers across devices. Respectively, we assign $5\%$, $35\%$, and $60\%$ of epochs to the warmup, the searching, and the fine-tuning phases. We apply architecture regularization after $15\%$ of epochs. However, in our experiments, the supernet is not excessively sensitive to changes in these hyperparameters.

\begin{table}
\begin{center}
\begin{tabular}{|c||cccc||cc|}
\hline
\textbf{Name} & \textbf{L} & \textbf{F} & \textbf{Exp} & \textbf{S} & \textbf{FLOPs} & \textbf{Params}\\
\hline\hline
\textbf{Small} & 10 & 16 & 3 & 1 & 57.7G & 3.2M\\
\textbf{-} & 14 & 16 & 6 & 1 & 109.4G & 10.7M\\
\textbf{Medium} & 14 & 64 & 6 & 1/4 & 380.1G & 22.3M\\
\textbf{Large} & 10 & 64 & 3 & 1 & 558G & 47.3M\\
\hline
\end{tabular}
\end{center}
\caption{Comparison of different models varying in size. See \cref{subsec:implementation} for reference.}
\label{table:model_sizes}
\end{table}

Following previous works \cite{autodeeplab,dcnas}, we use two optimization strategies to train architectural parameters and operation weights. For the former, we adopt Adam \cite{adam} with
a learning rate of 0.003 and weight decay of 0.001. For the latter, we use SGD parameterized by a learning rate of 0.003 and weight decay of 0.0005. As a data augmentation, we apply horizontal flipping, random scaling, color jittering, and random Gaussian noise. We set the batch size to 16 and train the network using crops of 512 ×1024. More details can be found in our implementation, which we release with the code.

We present different variants of models in \cref{table:model_sizes}. They vary in the number of layers \textit{L}, the filter multiplier \textit{F}, the expansion ratio \textit{Exp} in the inverted bottleneck, and the channel sampling ratio \textit{S}. Due to limited computational resources, we use the small model in the experiments concerning discretization error and entropy loss.

\subsection{Emergence of discretization error}\label{subsection:discretization_error}

\cref{fig:entropy} illustrates the average entropy of $c_\beta$. We can observe a sharp drop in entropy for the step scaling function, which can negatively impact training dynamics, especially at higher magnitudes. This observation aligns with our results presented in \cref{subsec:entropy_loss_exp}. The experiment also shows that the unregularized supernet, denoted by a blue line, has severely non-discretized architectural parameters, which could lead to the discretization error.

To study more thoroughly how discretization is impacted by entropy regularization, we gradually discretize supernets trained with different magnitudes of entropy loss. Specifically, we perform inference after dropping a certain number of redundant edges based on the strength of their architectural parameters. It is important to note that while this can effectively measure the relative importance of an edge within a cell, it might not optimally rank edges according to their relevance across different cells in different parts of the supernet. Nevertheless, we use it as a reasonable approximation.

The results are illustrated in \cref{fig:inference}. First, we observe a quick collapse of a standard unregularized DNAS supernet after dropping merely $20\%$ of its edges. Second, we empirically demonstrate a correlation between the strength of regularization and resistance to discretization. In particular, pruning $30\%$ of the transmissions in a heavily regularized supernet does not result in any noticeable drop in performance.

We conduct the same experiments, but this time they are followed with a short post-decoding fine-tuning, which adapts transferred weights to a discretized architecture. The aim is to investigate whether discretized architecture lies in the neighborhood of the optimal architecture in parameter space. In such a scenario, performing a relatively small number of updates could retrieve optimal performance. Our findings, presented in \cref{fig:inference_ft}, confirm that a strong entropy regularization can effectively address the discretization issue. Conversely, a vanilla DNAS supernet generates a qualitatively different architecture, partially explaining the low correlation observed by DCNAS~\cite{dcnas}.

\subsection{Negative impact of entropy loss}\label{subsec:entropy_loss_exp}

\begin{table}
\begin{center}
\begin{tabular}{|c||c|c|c|c|c|}
\hline
\textbf{Method} & \textbf{FLOPs (G)} & \textbf{s-mIoU} & \textbf{t-mIoU}\\
\hline\hline
Auto-DeepLab & 695 & 34.9\% & 80.3\%\\
DCNAS & 294.6 & 69.9\% & \textbf{81.2}\%\\
DCNAS (non-dense) & - & 51.7\% & 73.3\%\\
DPC & 684 & - & 80.9\%\\
Ours (Small) & 57.7 & 71.4\% & n/a \\
Ours (Medium) & 380.1 & 74.4\% & n/a \\
Ours (Large) & 558 & \textbf{75.3}\% & n/a \\
\hline
\end{tabular}
\end{center}
\caption{Comparison between different methods on the Cityscapes validation dataset. FLOPs are computed for the final networks and taken from DCNAS. In our case, we evaluate the performance of the supernet.}
\label{table:model_efficiency}
\end{table}

\setlength{\tabcolsep}{3pt}
\renewcommand{\arraystretch}{1.25}

As we highlight in \cref{section:introduction} and \cref{section:related_work}, the entropy loss term is an established solution to the discretization error in the literature. However, it causes a sudden drop in the entropy of architectural parameters, as illustrated in \cref{fig:entropy}. This might lead to a suboptimal convergence of the supernet. Dynamic regularization results in a more gradual entropy decrease.

We study the efficacy of the dynamic and static entropy losses on the Cityscapes validation set. Experiments are conducted with different scaling functions and magnitudes of the entropy loss ($c_\alpha$ and $c_\beta$). We report an average mIoU over three runs to obtain accurate estimates. Results are shown in \cref{table:entropy}. We observe a consistent drop in performance for higher entropy magnitudes, indicating the emergence of the discretization-exploration trade-off. A linearly scaled entropy loss term alleviates the issue and outperforms the default regularization technique, albeit converges poorly for higher entropy magnitude, emphasizing the necessity for a more robust approach.

\begin{table}
\begin{center}
\begin{tabular}{|c||ccc||c|}
\hline
\textbf{Method} & \textbf{Val} & \textbf{Coarse}& \textbf{ImageNet} & \textbf{Results}\\
\hline\hline
GridNet \cite{gridnet} & & & & 69.5\\
FRRN-B \cite{frrnet} & & & & 71.8\\
Ours (Large) & & & & 74.0\\
DCNAS \cite{dcnas} & & & & 82.8\\
\hline\hline
PSPNet \cite{pspnet} & \cmark & \cmark & \cmark & 81.2\\
DeepLabv3+ \cite{deeplabv3+} & \cmark & \cmark & \cmark & 81.3\\
HRNetV2 + OCR \cite{ocr} & \cmark & & \cmark & 83.9\\
\hline\hline
SparseMask \cite{sparsemask} & & & \cmark & 68.6\\
CAS \cite{cas} & \cmark & \cmark & \cmark & 72.3\\
GAS \cite{gas} & \cmark & \cmark & \cmark & 73.5\\
Auto-DeepLab \cite{autodeeplab} & \cmark & & & 80.4\\
Auto-DeepLab \cite{autodeeplab} & \cmark & \cmark & & 82.1\\
RSPNet \cite{rspnet} & \cmark & & \cmark & 81.4\\
DPC \cite{dpc} & \cmark & \cmark & \cmark & 82.7\\
DCNAS \cite{dcnas} & \cmark & \cmark & & 83.6\\
DCNAS + ASPP \cite{dcnas} & \cmark & \cmark & & 84.3\\
\hline
\end{tabular}
\end{center}
\caption{Cityscapes test set results. \textbf{Val}: Models are also trained using annotations from the validation set. \textbf{Coarse}: Models exploit coarse annotations. \textbf{ImageNet}: Models pretrained on ImageNet.}
\label{table:test_set}
\end{table}

\subsection{Proxyless searching}

We validate the single-stage searching protocol, our remedy for the discretization issue, in \cref{table:model_efficiency}. Namely, we report the mIoU of our approach and the state-of-the-art DNAS methods on the validation set. For all models, we also provide the number of floating-point operations. Our medium and large models outperform supernets of Auto-DeepLab and DCNAS in the searching stage by achieving $74.4\%$ and $75.3\%$, respectively. The large model matches Auto-DeepLab and DPC in the number of floating-point operations, whereas the medium network requires as little as $30\%$ more than DCNAS.

Importantly, the medium model attains $1.1\%$ higher mIoU compared to the non-dense variant of DCNAS, which is most comparable to ours in terms of the search space design. Given these results and the reasonable computational cost, the supernet trained in a proxyless way can be considered a viable drop-in replacement for the final network of the state-of-the-art DNAS algorithms. We hypothesize that incorporating long-range connections to the search space might further elevate the performance.

\begin{table}
\begin{center}
\begin{tabular}{|c||c|c|c|}
\hline
\multirow{2}{*}{\textbf{Method}} & \multicolumn{2}{|c|}{\textbf{Searching}} & \textbf{Retraining}\\
\cline{2-4}
& \textbf{Epochs} & \textbf{GPU (days)} & \textbf{Epochs}\\
\hline
Auto-DeepLab & 40 & 3 & 16000\\
DCNAS & 120 & 5.6 & 800\\
DPC & $28\text{k}\times80$ & 2600 & 500\\
Ours (Large) & 600 & 5.5 & -\\
\hline
\end{tabular}
\end{center}
\caption{Comparison of DNAS methods on Cityscapes in time efficiency. The searching stage time for Auto-DeepLab and DPC is provided for P100, which considerably overstates the costs. We estimate DPC epochs based on values of its hyperparameters \cite{dpc}.}
\label{table:time_comparison}
\end{table}

\setlength{\tabcolsep}{5pt}
\renewcommand{\arraystretch}{1.3}
\begin{table}
\begin{center}
\begin{tabular}{|c|c|c|c|c|}
\hline
\multicolumn{2}{|c|}{\textbf{Weights}} & \multicolumn{2}{c|}{\textbf{Architecture}}  & \multirow{2}{*}{\textbf{mIoU}}\\\cline{1-4}
\textbf{Fine} & \textbf{Coarse} & \textbf{Fine} & \textbf{Coarse} &\\
\hline\hline
0.5 & 0 & 0.5 & 0 & 69.9\% \\
1 & 0 & 1 & 0 & \textbf{75.3}\% \\
1 & 0 & 0 & 1 & 74.6\% \\
0 & 1 & 1 & 0 & 61.8\% \\
1 & 1 & 1 & 1 & 75.2\% \\
\hline
\end{tabular}
\end{center}
\caption{Results of the large model trained with different data splits on the validation set. \textbf{0}: Annotations are not used in training. \textbf{0.5}: Half of the annotations are used exclusively to optimize given parameters. \textbf{1}: All of the annotations are used. In case we use both fine and coarse annotations, we train the parameters using all the data and then fine-tune them with only fine labels.}
\label{table:dataset_split}
\end{table}

We present results for the test set in \cref{table:test_set}. However, we do not further optimize performance. We employ the same training and inference procedures as used for the validation set, which may potentially understate the obtained results.

Regarding training time, we benchmark our approach against other DNAS works in \cref{table:time_comparison}. In the conventional searching-retraining procedure, only a small amount of time is dedicated to finding the optimal architecture~\cite{darts,dpc,autodeeplab,dcnas}. The bulk of computational resources are devoted to retraining the derived architecture, significantly increasing the costs of using other DNAS methods. The single-stage searching protocol eliminates the need for a time-consuming retraining stage, thus providing the optimal architecture in only 5.5 GPU days.

\subsection{Optimal dataset split}
DNAS methods commonly address the emerging bilevel optimization problem by training architectural parameters, $\{\alpha, \beta\}$, and operation weights, $\{\omega\}$, on two disjoint subsets of the training dataset. DARTS introduced this to prevent poor architecture generalization. MiLeNAS~\cite{milenas} argues that optimizing architectural parameters on the entire training dataset is optimal. Results from experiments with varied splits are presented in \cref{table:dataset_split}. We show that joint optimization using the same data yields the best results, provided that batches are sampled separately in each iteration for both sets of parameters. We also use two separate optimizers. Combining fine with coarse annotations does not offer significant benefits, potentially due to the supernet's undertraining. Surprisingly, performing architecture updates with batches consisting exclusively of coarse annotations matches the training performance obtained using other splits. Additionally, our supernet achieves superior results in the searching stage, even when using a less optimal data split than the one used by DCNAS.

\begin{figure*}
  \centering
  \includegraphics[scale=0.6]{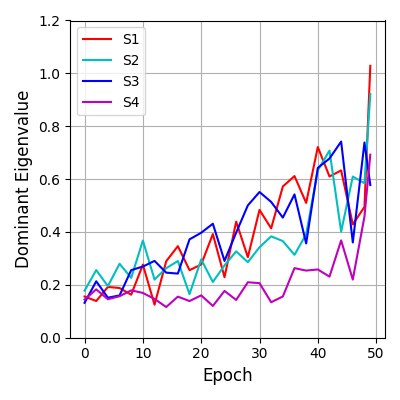}
  \includegraphics[scale=0.6]{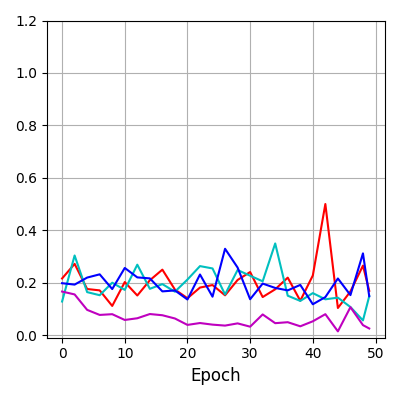}
  \caption{
  (left) dominant eigenvalues of $\nabla_{\alpha}^2\mathcal{L}_{valid}$ on four different search spaces with a dataset split; (right) dominant eigenvalues when searching on a single dataset. All experiments were conducted on CIFAR 10 dataset.
  }
  \label{fig:eigenvalues}
\end{figure*}

To further verify our findings, we test if the new splitless dataset procedure affects architecture degeneration in DARTS~\cite{darts}. RobustDARTS~\cite{robustdarts} showed that dominant eigenvalue of $\nabla_{\alpha}^2\mathcal{L}_{valid}$ is significantly increasing during searching and that there exists a correlation between large dominant eigenvalue and architecture degeneration across four different search spaces S1, S2, S3 and S4. First, we reproduce experiments presented in the paper for standard DARTS on CIFAR-10. Results are illustrated in \cref{fig:eigenvalues} (left). We indeed observe that in all four cases dominant eigenvalues are steadily increasing and reach much larger values at the end of the searching. Second, we perform the same experiments, but with parameters optimized on the union of training and validation dataset. \cref{fig:eigenvalues} (right) shows that this time in all four search spaces dominant eigenvalues are kept relatively small, which suggests that the network doesn't end up in a sharper local minima that would cause an architecture degeneration.

\section{Conclusions}

In this work, we conduct comprehensive experiments to investigate the discretization error. Our findings indicate that this issue emerges in an unregularized supernet during the searching stage. We study an established method for addressing this issue through entropy architecture regularization and highlight its shortcomings. As a remedy, we propose the single-stage searching protocol, a novel way of finding optimal neural networks using a gradient-based technique. Our method robustifies the DNAS framework by eliminating the discretization error that other methods suffer from. We show that it matches other state-of-the-art approaches in terms of performance and results within a similar search space. Also, we demonstrate that the joint optimization of architecture and weights on the full dataset yields better results, and prevents a well-known architecture degeneration phenomenon in DARTS. We believe that our work can be a starting point for creating even more powerful DNAS models. In future work, we aim to efficiently incorporate long-range connections into the search space to improve results even further.

\subsection*{Acknowledgements}

We would like to express our gratitude to Nvidia and Google’s TPU Research Cloud for making their compute resources available to us. Similarly, we would like to thank PLGrid and ICM UW for sharing their GPU clusters. Neural Architecture Search is one of the most compute-intensive areas of deep learning, and this work could not have been possible without immense amounts of computing power.
{
    \small
    \bibliographystyle{ieeenat_fullname}
    \bibliography{main}
}

\end{document}